\documentclass[letterpaper, 10 pt, conference]{IEEEtran} 
\usepackage{wrapfig}
\usepackage{float}
\usepackage[tbtags]{amsmath}
\usepackage{amssymb,verbatim,bm,color,times,mathtools}
\usepackage[font=small,format=plain]{caption}
\usepackage{stfloats}
\usepackage{subcaption}
\usepackage{graphicx}
\usepackage{blindtext}
\usepackage{scrextend}
\usepackage{siunitx}
\usepackage{psfrag}
\usepackage{multicol}
\usepackage{pgfplots}
\usepackage{ulem}
\usepackage{tikz}
\usetikzlibrary{automata, positioning}
\usepackage{xspace}
\usepackage{tabularx}
\usepackage{booktabs}
\usepackage{enumitem}

\usepackage{amsmath}

\DeclareFontFamily{U}{mathb}{\hyphenchar\font45}
\DeclareFontShape{U}{mathb}{m}{n}{
	<5> <6> <7> <8> <9> <10> gen * mathb
	<10.95> mathb10 <12> <14.4> <17.28> <20.74> <24.88> mathb12
}{}
\DeclareSymbolFont{mathb}{U}{mathb}{m}{n}
\DeclareFontSubstitution{U}{mathb}{m}{n}

\let\dot\relax
\DeclareMathAccent{\dot}{0}{mathb}{"39}
\let\ddot\relax
\DeclareMathAccent{\ddot}{0}{mathb}{"3A}
\let\dddot\relax
\DeclareMathAccent{\dddot}{0}{mathb}{"3B}
\let\ddddot\relax
\DeclareMathAccent{\ddddot}{0}{mathb}{"3C}

\usepackage{nomencl}
\makenomenclature

\RequirePackage{ifthen}
\renewcommand\nomgroup[1]{%
  \item[\bfseries
  \ifthenelse{\equal{#1}{D}}{\item[\textbf{Subscripts}]}{%
  \ifthenelse{\equal{#1}{E}}{\item[\textbf{Superscripts}]}{%
  \ifthenelse{\equal{#1}{B}}{\item[\textbf{Symbols}]}{%
  \ifthenelse{\equal{#1}{A}}{\item[\textbf{Acronyms}]}{%
  \ifthenelse{\equal{#1}{C}}{\item[\textbf{Operators}]}{%
  \ifthenelse{\equal{#1}{F}}{\item[\textbf{Notations}]}{}}}}}}]
}

\usepackage{cite}
\usepackage{algorithm}
\usepackage{algpseudocode}
\usepackage{siunitx}

\begin{document}
\title{Probabilistic Latent Variable Modeling for \\ Dynamic Friction Identification and Estimation}
\author{Victor Vantilborgh\textsuperscript{1,2,*}, Sander De Witte\textsuperscript{1,2}, Frederik Ostyn\textsuperscript{1,2}, Tom Lefebvre\textsuperscript{1,2} and Guillaume Crevecoeur\textsuperscript{1,2}
\thanks{\textsuperscript{1,2}V.V., S.D.W., F.O., T.L., and G.C. are with Dynamic Design Lab (D\textsuperscript{2}LAB) from the Department of Electromechanical, Systems and Metal Engineering, Ghent University, B-9052 Ghent, Belgium e-mail: \{surname.name\}@ugent.be.}%
\thanks{\textsuperscript{2}V.V., S.D.W., F.O., T.L., and G.C. are members of Core Lab MIRO, Flanders Make, the strategic research center for the manufacturing industry in Flanders, Belgium.}%
\thanks{*Corresponding author.}
\vspace{-20pt}
}
    
    
\maketitle

\begin{abstract}
Precise identification of dynamic models in robotics is essential to support dynamic simulations, control design, friction compensation, output torque estimation, etc. A longstanding challenge remains in the development and identification of friction models for robotic joints, given the numerous physical phenomena affecting the underlying friction dynamics which result into nonlinear characteristics and hysteresis behaviour in particular. These phenomena proof difficult to be modelled and captured accurately using physical analogies alone. This has motivated researchers to shift from physics-based to data-driven models. Currently, these methods are still limited in their ability to generalize effectively to typical industrial robot deployement, characterized by high- and low-velocity operations and frequent direction reversals. Empirical observations motivate the use of dynamic friction models but these remain particulary challenging to establish. To address the current limitations, we propose to account for unidentified dynamics in the robot joints using latent dynamic states. The friction model may then utilize both the dynamic robot state and additional information encoded in the latent state to evaluate the friction torque. We cast this stochastic and partially unsupervised identification problem as a standard probabilistic representation learning problem. In this work both the friction model and latent state dynamics are parametrized as neural networks and are integrated in the conventional lumped parameter dynamic robot model. The complete dynamics model is directly learned from the noisy encoder measurements in the robot joints. We use the Expectation-Maximisation (EM) algorithm to find a Maximum Likelihood Estimate (MLE) of the model parameters. The effectiveness of the proposed method is validated in terms of open-loop prediction accuracy in comparison with baseline methods, using the Kuka KR6 R700 as a test platform.
\end{abstract}


\section{Introduction}
\label{sec:introduction}
Precise identification and modeling of dynamic behavior hold significant potential to improve performance of robotic systems in multiple aspects. It is essential for achieving accurate dynamic simulations and improve performance in motion control tasks. Model-based control schemes take into account all modelled phenomena and the effects they introduce on the system dynamics. Among these, friction is one of the most dominant and undesired phenomena. Friction arises in the joints where relative motion occurs between contact surfaces, resulting in energy dissipation. This highly nonlinear phenomenon depends on several factors, such as surface material, type of lubricant, joint speed, temperature, axial load and so forth \cite{kim2019}. Including all friction characteristics in the model is therefore exhaustive and very challenging.

It is common to assume a static friction model taking into account Coulomb and viscous friction, as well as the Stribeck effect \cite{stribeck}. However, typical motion of robotic systems in an industrial environment is characterized by many direction reversals and high and low velocity operations, resulting into dynamic and hysteretic friction behaviour that static models fail to capture adequately. Dynamic friction models aim to adress this issue. The Lugre model \cite{Lugre} describes the transition from static to dynamic friction by introducing a dynamic latent state variable that represents the average deflection of the bristles at the contact surfaces in the joint. The Generalized Maxwell-Slip (GMS) model \cite{GMS} adresses the hysteris behaviour in the presliding regime, dominated by adhesive forces, which has implications around velocity reversals. Other factors such as backlash, elastic deformations, microscopic interactions between the contact surfaces, varying load etc. are often neglected. Current research efforts typically focus on incorporating one or more of these effects in a friction model \cite{iskandar2019, weigand2021, tadese2021}. However, combining all of these into a single model would require significant engineering effort and result into highly complex models, also impeding efficient identification.

Recently, promising results have been obtained using Deep Learning (DL) methods that directly model friction as an input-output model from encoder and torque measurements. In \cite{tu2019, liu2021},  Neural Networks (NNs) are used to learn a static friction characteristic. Specialized architectures for friction modelling have been introduced, e.g. by \cite{guo2019}, that propose ‘jumping sigmoid’ activation functions for modelling the discontinuous friction characteristic. Additionally, research has focused on the inclusion of various factors influencing friction into data-driven models, with temperature and load torque being the most commonly addressed effects \cite{simoni2015, visiolo2017, carlson2015, gao2017}. A minority of the work aims at establishing dynamic models. In \cite{scholl2024}, a hybrid approach of extending the conventional dynamic Lugre model with a static neural network, that acts as residual term for correcting erros of the Lugre model. In \cite{hirose2017} RNN and LSTM are used to obtain a dynamic friction model. However, these fully data-driven dynamic friction models require large datasets and have not demonstrated consistent performance across varying velocities, direction reversals, different loads, and high degree-of-freedom (DOF) robotic systems. Moreover, they can lead to non-Markovian state-space. Futhermore, RNNs and LSTMs internalize the latent state estimation which impedes their general applicability post identification.

This work is motivated by the observation that friction in robotic systems is an intrinscally dynamic phenomenon. Physics-based models fail to capture all effects succesfully due to their prescribed structure. Supervised DL methods have shown their potential in capturing complicated input-output relation without structural bias however they are not straightforward to apply in the partially unsupervised setting of latent state dynamics.  

To address these problems, we propose the following contributions
\begin{enumerate}
    \item We describe the dynamic model of the robotic system as a Probabilistic State-Space Model (PSSM). The unknown friction torque is parametrized by a NN. The friction dynamics are modelled by means of latent variables that represent the (partially) unknown underlying state of the system. The friction torque and dynamics are identified jointly with the conventional lumped parameter model.
    \item Through the Expectation-Maximization (EM) algorithm and Sequential Monte Carlo (SMC) techniques we obtain a Maximum Likelihood Estimate (MLE) of the PSSM. Our identification method produces an accurate model and does not necessitate pre-processing or noise handling of the sensor data.
    \item  We evaluate our approach on a KUKA KR6 700 industrial robot and show the improved results compared to the existing literature.
\end{enumerate}

\section{Background}\label{sec:background}
\subsection{Robot dynamics}
The dynamics of kinematic chains composed of rigid bodies are typically formulated using either the Newton-Euler or Lagrangian methods, yielding a set of equations generally referred to as the inverse dynamic model
\begin{equation}\label{eq:IDIM}
    \mathbf{M(q)} \ddot{\mathbf{q}} + \mathbf{c}(\mathbf{q}, \dot{\mathbf{q}}) + \mathbf{g(q)} + \boldsymbol{\tau}_f(\mathbf{q}, \dot{\mathbf{q}}) = \boldsymbol{\tau}_m
\end{equation}
Here $\mathbf{q}$, $\dot{\mathbf{q}}$ and $\ddot{\mathbf{q}}$ denote the joint position, velocity and acceleration, respectively. Here, $\mathbf{M(q)}$ represents the positive definite inertia matrix, $\mathbf{c}(\mathbf{q}, \dot{\mathbf{q}})$ accounts for the Coriolis and centrifugal effects, $\mathbf{g(q)}$ denotes the gravitational torque, and $\boldsymbol{\tau}_f$ and $\boldsymbol{\tau}_m$ describe the friction and motor torque, respectively. Friction is typically modeled as a function depending on the joint velocity $\dot{\mathbf{q}}$ and sometimes also the position $\mathbf{q}$.

\subsection{Conventional identification method}
Conventional methods for robotic system identification are typically based on the inverse dynamics model \cite{deschutter2007, leboutet2021}. The standard dynamic robot model is linear-in-the-parameters so that it is possible to rewrite (\ref{eq:IDIM}) as follows \cite{gautier1990}
\begin{equation}\label{eq:linear}
    \boldsymbol{\tau}_m = Y(\mathbf{q}, \dot{\mathbf{q}}, \ddot{\mathbf{q}}) \boldsymbol{\Theta}
\end{equation}
Here $Y(\mathbf{q}, \dot{\mathbf{q}}, \ddot{\mathbf{q}})$ is a regressor matrix and $\boldsymbol{\Theta}$ the vector of standard inertial lumped parameters. For rigid bodies these standard inertial parameters include the moment of inertia tensor elements, $\{I_{xx, j}, I_{yy, j}, I_{zz, j}, I_{xy, j}, I_{xz, j}, I_{yz, j}\}$, the centre of mass, $\{r_{x,j}, r_{y,j},r_{z,j}\}$, the mass, $m_j$, and the parameters describing the friction law, $\boldsymbol{\theta}_{f,j}$, for each link and joint $j$. For example, a simple model where only Coulomb and viscous friction are considered -- modeled as a linear function of the joint velocity and its sign, i.e. $\tau_{f,j} = \nu_{c,j} \mathrm{sign}(\dot{q}_j) + \nu_{v,j} \dot{q}_j$, -- the parameter $\boldsymbol{\theta}_{f,j}$ would represent the Coulomb and viscous friction coefficients, $\{\nu_{c,j},\nu_{v,j}\}$. 

The base inertial parameters are the minimal set of identifiable parameters to parametrize the dynamic model and can be obtained through proper regrouping of the linear system (\ref{eq:linear}) by means of linear relations or a numerical method \cite{gautier1990}. This overdetermined linear system can be solved by Least Squares (LS), after collecting a qualitative dataset. Therefore, periodic excitation reference trajectories of $T+1$ timesteps along, $\{\mathbf{q_{0:T}}, \dot{\mathbf{q}}_{0:T}, \ddot{\mathbf{q}}_{0:T}\}$, are designed to persistently excite the base inertial parameters at a user defined sampling rate. By controlling the frequency spectrum of the periodic excitation signal, the signal-to-noise level can be improved through exact frequency domain post-processing of the data. 

Extending this linear-in-the-parameters robot model to include more advanced friction models further improves accuracy. Such friction models are, however, nonlinear-in-the-parameters and considerably complicate the parameter estimation. This nonlinear optimization problem is typically solved using iterative gradient-based methods. For a physics-based model, such as the Stribeck model \cite{stribeck}, the values obtained for the linear friction parameters can serve as inspiration for the initial guess of related parameters in the nonlinear model, with optimization carried out by, for example, the Downhill-Simplex method \cite{simplex}.  

For data-driven models, such as a neural networks, the lack of a physically inspired model structure and the high number of model parameters make it impossible to determine a meaningful initial guess. These models are optimized using speciliazed stochastic gradient-based optimization methods, which often require intensive hyperparametertuning. 

Things complicate further for dynamic friction models, such as the Lugre model and the GMS model. Here the supervised learning structure breaks down, due to the presence of one or more unobserved dynamic variables for which labeled data is unavailable. These unobserved variables must be accounted for, and the parameters describing them need to be inferred implicitly from the measurements of $\mathbf{q}$ and $\dot{\mathbf{q}}$. There is no way of assessing the validity of the proposed physical structure for the unobserved dynamics or detecting potential mismatches between the model's assumed structure and the actual latent dynamics affecting the robotic system, other than evaluating its contribution to the model structure.

Moreover, as the model structure becomes more complex, the quality of the data which is used for identification becomes increasingly more important and should be obtained in such manner as to contain the effects of the different friction regimes. From this it is clear that there is an important trade-off between model accuracy and the overall modeling and estimation complexity.

\label{sec:classicID}
\section{Methodology}\label{sec:methods}
Our goal is to identify a dynamic robot model based on $N$ data sequences $\mathcal{D} = \{\mathbf{y}_{0:T}^{n}, \mathbf{u}_{0:T}^{n}\}_{n=1}^{N}$, with $\mathbf{y}=\{\mathbf{q},\dot{\mathbf{q}}\}$ and $\mathbf{u}=\boldsymbol{\tau}_m$. We further desire the model to have a similar structure as the model in (\ref{eq:IDIM}) but additionaly incorporate latent variables to account for any unidentified friction dynamics. This implies that we can introduce a memory variable into the friction torque evaluation but also need to come up with a dynamic model for the latent variables. To this end we establish a prestructured Probabilistic State Space Model (PSSM) that incoporates the standard dynamic model with arbritrary friction torque and extends it with the latent dynamics. Unlike traditional dynamic friction models, which impose predefined dynamic states based on physical assumptions (e.g. the bristle deflection in the Lugre model), our approach does not enforce strong priors on these latent states. Both the friction torque models as well as the latent state dynamics are represented by NNs. Based on the PSSM framework we can rely on existing techniques to identify the lumped parameters as well as the NN parameters simultaneously.

\subsection{Probabilistic State Space Models}
A PSSM is characterised by an initial, $p_{\boldsymbol{\theta}}(\mathbf{x}_0)$, transition $p_{\boldsymbol{\theta}}(\mathbf{x}_t|\mathbf{x}_{t-1},\mathbf{u}_{t-1})$, and, emission density, $p_{\boldsymbol{\theta}}(\mathbf{y}_t|\mathbf{x}_t)$. We assume that the state, $\mathbf{x}_t\in\mathcal{X}\subset\mathbb{R}^{n_x}$, input $\mathbf{u}_t\in\mathcal{U}\subset\mathbb{R}^{n_u}$ and $\mathbf{y}_t\in\mathcal{Y}\subset\mathbb{R}^{n_y}$ are continuous vector quantities. Here, $t \in \mathbb{N}$ is the discrete time index. Furthermore, we impose the Markov assumption. Finally we assume that the densities constituting the PSSM are parametrised by some parameter $\boldsymbol{\theta} \in \boldsymbol{\Theta} \subset \mathbb{R}^{n_\theta}$. The goal is to find a representation of these densities and a value for the parameter $\boldsymbol{\theta}$ that describes them.

Based on the former modelling assumptions, the joint probability over the measurement and state trajectory can be decomposed as follows 
\begin{equation}\label{eq:integrand}
\begin{multlined}
p_{\boldsymbol{\theta}}(\mathbf{y}_{0:T}, \mathbf{x}_{0:T} | \mathbf{u}_{0:T})  \\
\begin{aligned}
    &= p_{\boldsymbol{\theta}}(\mathbf{y}_{0:T}|\mathbf{x}_{0:T})p_{\boldsymbol{\theta}}(\mathbf{x}_{0:T}|\mathbf{u}_{0:T}) \\
    &= p_{\boldsymbol{\theta}}(\mathbf{x}_0)p_{\boldsymbol{\theta}}(\mathbf{y}_0|\mathbf{x}_0) \prod_{t=1}^{T} p_{\boldsymbol{\theta}}(\mathbf{y}_t|\mathbf{x}_t) p_{\boldsymbol{\theta}}(\mathbf{x}_t | \mathbf{x}_{t-1}, \mathbf{u}_{t-1})
\end{aligned}
\end{multlined}
\end{equation}

In this work, we impose a specific structure on the transition function by reformulating the inverse dynamic model (\ref{eq:IDIM}) as a state-space model and extending the robot state, $\mathbf{x}_t$, with latent states $\mathbf{z}_t$
\begin{equation}
\begin{aligned}
    \dot{\mathbf{x}}_t = \begin{bmatrix} \ddot{\mathbf{q}}_t \\ \dot{\mathbf{q}}_t \\ \dot{\mathbf{z}}_t\end{bmatrix} &= \begin{bmatrix} \mathbf{M(q)}^{-1} (\boldsymbol{\tau}_m - \mathbf{c}(\mathbf{q}, \dot{\mathbf{q}}) - \mathbf{g(q)} - \boldsymbol{\tau}_{f, \boldsymbol{\theta}}(\mathbf{x_t})) \\ \dot{\mathbf{q}}_t \\ \boldsymbol{\eta}_{\boldsymbol{\theta}}(\mathbf{x_t})\end{bmatrix} \\ &= \mathbf{f}'_{\boldsymbol{\theta}}(\mathbf{x_t}, \boldsymbol{\tau}_m)
\end{aligned}
\end{equation}
where $\boldsymbol{\eta}_{\boldsymbol{\theta}}(\mathbf{x_t})$ defines the latent variable dynamics.

Numerical integration methods can be used to solve this derivative function over the sampling interval to compute the next state, such that $\mathbf{x}_{t} = \mathbf{f}_{\boldsymbol{\theta}}(\mathbf{x}_{t-1}, \mathbf{u}_{t-1})$. A more general Probabilistic State-Space Model can now be formulated as
\begin{equation}
\begin{aligned}
    \label{eq:PSSM}
    p(\mathbf{x}_{t}| \mathbf{x}_{t-1},\mathbf{u}_{t-1}) &= \mathbf{f}_{\boldsymbol{\theta}}(\mathbf{x}_{t-1},\mathbf{u}_{t-1}) + \mathbf{w}_t, ~\mathbf{w}_t \sim \mathcal{N}_{\boldsymbol{\theta}}(0,\mathrm{Q})\\
    p(\mathbf{y}_t|\mathbf{x}_t) &= \mathbf{g}_{\boldsymbol{\theta}}(\mathbf{x}_t) + \mathbf{v}_t, ~\mathbf{v}_t \sim \mathcal{N}_{\boldsymbol{\theta}}(0,\mathrm{R})
\end{aligned}
\end{equation}
with $\mathbf{g}_{\boldsymbol{\theta}}(\mathbf{x}_t)$ the measurement model. The random variables $\mathbf{w}_t$ and $\mathbf{v}_t$ serve as sources of process and measurement noise, respectively. The covariance matrices in (\ref{eq:PSSM}) can be included in $\boldsymbol{\theta}$.

Parameterizing the transition function $\boldsymbol{\eta}_{\boldsymbol{\theta}}(\mathbf{x}_t)$ of the latent states by a neural network enables the learning of highly nonlinear dependencies. Furthermore, parameterizing the friction characteristics $\boldsymbol{\tau}_{f, \boldsymbol{\theta}}(\mathbf{x_t})$ and linking them to the complete extended state $\mathbf{x}_t$ provides the necessary representational flexibility to obtain accurate dynamics of the robotic system. The model parameters, $\boldsymbol{\theta}$, then consist of the parameters of these two neural networks and the remaining base inertial parameters.

\subsection{Identification method}
\subsubsection{Maximum Likelihood Estimation}
The identification of the PSSM (\ref{eq:PSSM}) is achieved by determining the Maximum Likelihood Estimate (MLE) of the model parameters $\boldsymbol{\theta}$ that maximizes the marginal likelihood of the observed data $\{\mathbf{y}_{0:T}^{n}\}$, 
\begin{equation}
    \label{eq:MLE}
    \hat{\boldsymbol{\theta}}_{\text{MLE}} = \max_{\boldsymbol{\theta}} \mathcal{L}(\{\textbf{y}^{n}_{0:T}\}_{n=1}^N) = \max_{\boldsymbol{\theta}} \sum_{n=1}^N \log p_{\boldsymbol{\theta}}(\textbf{y}_{0:T}^n | \textbf{u}_{0:T}^n)
\end{equation}

For a PSSM, this likelihood, for a single data sequence, can be computed via the integral
\begin{equation}
    p_{\boldsymbol{\theta}}(\mathbf{y}_{0:T}|\mathbf{u}_{0:T}) = \int p_{\boldsymbol{\theta}}(\mathbf{y}_{0:T},\mathbf{x}_{0:T}|\mathbf{u}_{0:T}) d\mathbf{x}_{0:T} 
\end{equation}
A tractable expression for this integral can be obtained by substitution of (\ref{eq:integrand}) in the integrand. It is, however, well recognized that optimizing the Maximum Likelihood objective (\ref{eq:MLE}) presents significant challenges, given that $\mathbf{x}_{0:T}$ is (partially) unobserved and $p_{\boldsymbol{\theta}}(\mathbf{x}_t | \mathbf{x}_{t-1}, \mathbf{u}_{t-1})$ remains unknown prior to the estimation of the parameter $\boldsymbol{\theta}$. We make use of the Expectation-Maximization (EM) algorithm \cite{krishnan} to adress these challenges. 

The EM-algorithm is a two-step iterative optimization procedure. The Expectation step deals with $\mathbf{x}_{0:T}$ being unavailable by assuming a value $\boldsymbol{\theta}^*$ on the model parameters $\boldsymbol{\theta}$. This allows to evaluate the model structure (\ref{eq:PSSM}) and estimate the "missing" data $\mathbf{x}_{0:T}$. Given this assumption and the observed data $\mathbf{y}_{0:T}$, the data likelihood function $\mathcal{L}$ can then be approximated by its minimum variance estimate, $\mathcal{Q}_{\boldsymbol{\theta},\boldsymbol{\theta}^*}$, also known as the Evidence Lower Bound (ELBO):
\begin{equation}
    \label{eq:Q}
    \begin{aligned}
    \mathcal{Q}_{\boldsymbol{\theta}, \boldsymbol{\theta}^*} &= \mathbb{E}_{\boldsymbol{\theta}^*} \left[ \log p_{\boldsymbol{\theta}}(\mathbf{y}_{0:T},\mathbf{x}_{0:T}|\mathbf{u}_{0:T})|\mathbf{y}_{0:T} \right] \\
    &= \int \log p_{\boldsymbol{\theta}}(\mathbf{y}_{0:T},\mathbf{x}_{0:T}|\mathbf{u}_{0:T}) p_{\boldsymbol{\theta}^*}(\mathbf{x}_{0:T}|\mathbf{u}_{0:T},\mathbf{y}_{0:T}) d\mathbf{x}_{0:T}
    \end{aligned}
\end{equation}
Next, in the Maximization-step, the functional $\mathcal{Q}_{\boldsymbol{\theta},\boldsymbol{\theta}^*}$ is optimized for $\boldsymbol{\theta}$, producing an updated estimate for $\boldsymbol{\theta}^*$. This procedure is repeated until convergence. The EM-algorithm is summarized in Algorithm 1.
\begin{algorithm}[b]
    \caption{Expectation-Maximization Algorithm}
    \begin{algorithmic}[1]\label{alg:EM}
    \State Set $k=0$, initialize $\boldsymbol{\theta}_0$
    \Repeat
        \State E-step: calculate $\mathcal{Q}(\boldsymbol{\theta}, \boldsymbol{\theta}_k)$
        \State M-step: $\boldsymbol{\theta}_{k+1} = \arg\max_{\boldsymbol{\theta}} \mathcal{Q}(\boldsymbol{\theta}, \boldsymbol{\theta}_k)$
    \Until{convergence: $\mathcal{Q}(\boldsymbol{\theta}_{k}, \boldsymbol{\theta}_{k-1}) - \mathcal{Q}(\boldsymbol{\theta}_{k-1}, \boldsymbol{\theta}_{k-2}) \tiny \rightarrow 0$}
    \end{algorithmic}
\end{algorithm}

An additional challenge is the explicit evaluation of these Bayesian integrals, for which there is typically no hope of finding an analytical solution in the general nonlinear case. We resort to Sequential Monte Carlo (SMC) methods as emperical approximations of these Bayesian integrals.

\subsubsection{Sequential Monte Carlo}
The computation of $\mathcal{Q}_{\boldsymbol{\theta},\boldsymbol{\theta}^*}$ primarly depends on the smoothed density $p_{\boldsymbol{\theta}^*}(\mathbf{x}_{0:T}|\mathbf{u}_{0:T},\mathbf{y}_{0:T})$, which can typically only be calculated once the filtered distribution $p_{\boldsymbol{\theta}^*}(\mathbf{x}_{t}|\mathbf{u}_{0:t},\mathbf{y}_{0:t})$ is available, and expectations with respect to it. In this paper, we numerically approximate these posterior probabilities by relying on SMC methods, more specifically Sequential Importance Resampling (SIR) methods, which are better known under the informal title of \textit{particle filters and smoothers}.

The fundamental idea underlying SIR methods is to approximate the integrals of the filtering and smoothing distribution by a sum of \textit{sufficiently many} ($N_p$) uncorrelated samples $\hat x^i$, i.e. the particles. This approximation is expressed as 
\begin{equation}
    \label{eq:smoother}
    p(\mathbf{x}_{t}|\mathbf{y}_{0:T}, \mathbf{u}_{0:T}) \approx \sum_{i=1}^{N_p} w_{t|0:T}^{i} \delta (\mathbf{x}_t - \hat{\mathbf{x}}_t^i)
\end{equation}
where $\delta$ is the Dirac delta operator and $w_{t|0:T}^{i}$ the weights that quantify the likelihood of the $i^\text{th}$ particle $\hat{ \mathbf{x}}_t^i$ at moment $t$ given the sequence of observations $\mathbf{y}_{0:T}$. We refer to \cite{sarkka} for an in-depth discussion.

We would like te note that, while the conventional identification method discussed in Section \ref{sec:classicID} is not applicable to the proposed latent variable model, the identification approach based on MLE can be employed for all the aforementioned friction models. This probabilistic framework inherently manages process and measurement noise, and it allows for the simultaneous estimation of state variables and model parameters within a unified algorithm. Upon convergence, the identification algorithm provides a useful byproduct in the form of a state estimator, which can be utilized online for various purposes. A counterargument is that the recursive computations are time-consuming and convergence speed is typically slower.

\section{Experimental validation}\label{sec:experiments}
\subsection{Test setup and dataset}
\begin{figure}[h]
    \centering
    \includegraphics[width=0.8\columnwidth]{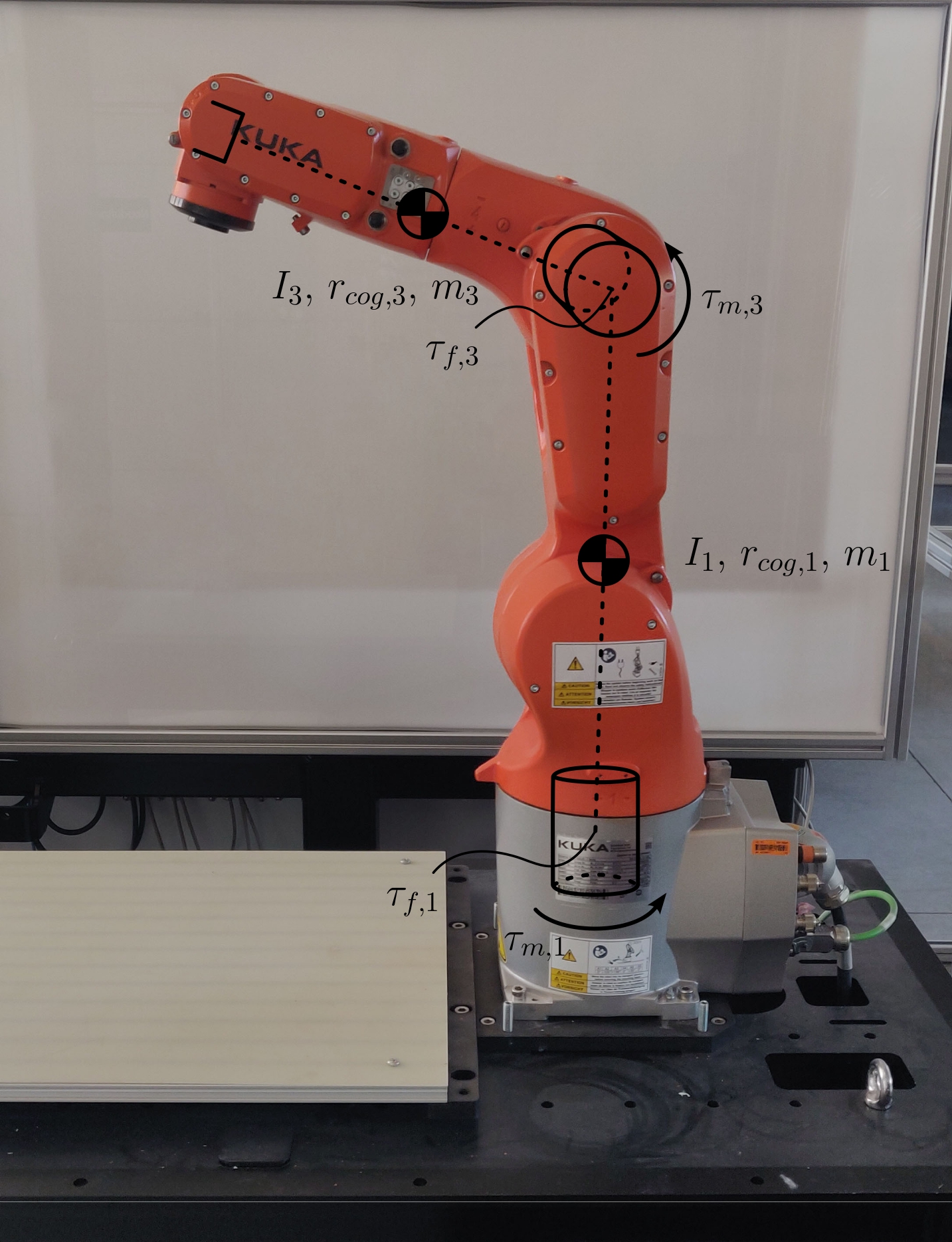}
    \caption{Experimental test setup with KUKA KR6 R700.}
    \label{fig:kuka}
\end{figure}

The position controlled KUKA KR6 R700 industrial robot is used as a validation platform. Fig. \ref{fig:kuka} depicts the experimental setup. A dataset is collected that aims to capture dynamic (friction) behaviour, therefore, the robot joints are excited simultaneously using varying velocity profiles that include direction changes. For each trajectory, the joint positions and velocties are collected, along with the motor torque, which are computed directly from the motor currents.
The dataset is collected according to the guidelines of the conventional identification method \cite{deschutter2007, leboutet2021} in order to allow proper camparison with the existing methods, as discussed in section \ref{sec:classicID}. The training dataset is designed as 3 different trajectories, each executed twice, with duration of 31.4s each, resulting in only about 3 minutes of training data. The validation dataset consist of a fourth trajectory of the same length. Design of experiment for each of the trajectories is formulated as a optimization problem to incorporate all physical constraints such as position, velocity, accelaration, jerk and self collision. A joint position reference trajectory is parametrized as a sum of sines and cosines
\begin{equation}
    \mathbf{q}(t) = \sum_{k=1}^{K} \left(\frac{\mathbf{a}_k}{k \omega_n} \sin(k \omega_n t) - \frac{\mathbf{b}_k}{k \omega_n} \cos(k \omega_n t)\right)
\end{equation}
with $K=20$ coefficients and $\omega_n=\frac{1}{5}\text{Hz}$ the base frecuency. The KUKA Robot Sensor Interface (RSI) software was used for implementing and applying these cyclic excitation signals and collect the data at a sample rate of $\Delta t = 0.004s$.

In contrast to \cite{deschutter2007},  we do not optimize the experiment with respect to the condition number. Although this criterion greatly improves the measurement information quality, i.e. by ensuring that all model parameters are excited sufficiently, the condition number is still a model-dependent criterion. As this is a dataset for nonlinear black-box identification, we decided against using model knowledge for the design of experiments. The optimization problem is given by

\begin{equation}
 \begin{aligned}
      \{\mathbf{a}^*,\mathbf{b}^*\} &= \arg\min_{\mathbf{a}, \mathbf{b}} J(\mathbf{q}(t)) \\
      \text{s.t. } &[\mathbf{q}, \dot{\mathbf{q}}, \ddot{\mathbf{q}} ]_{min} < [\mathbf{q}, \dot{\mathbf{q}}, \ddot{\mathbf{q}}] < [\mathbf{q}, \dot{\mathbf{q}}, \ddot{\mathbf{q}}]_{max} \\
      & [\mathbf{q}(t), \dot{\mathbf{q}}(t), \ddot{\mathbf{q}}(t)] = 0 \hspace{20pt} \forall t \in \{0, T\}
 \end{aligned}
\end{equation}
We choose $J(\mathbf{q}) = 0$, in accordance with the findings in \cite{weigand2023} that the empty objective function captures both stationary and high-velocity effects well. The random sampling of the initial guesses for $\mathbf{a}$ and $\mathbf{b}$ leads to different signal outcomes.

\section{Results}
The proposed method was benchmarked against several well-established and state-of-the-art friction models, including a simple model (i.e. Coulomb and viscous friction), the Stribeck characteristic, the Lugre model, the GMS model, a fully connected neural network, and a RNN. A quantitative comparison of open-loop prediction performance is provided, along with a qualitative analysis of the identified friction characteristics.

\subsection{Implementation details}
Identification of the benchmark models from the noisy measurements using the conventional identification method may lead to bias in the resulting models. We make use of a non-causal zero-phase digital filter with flat amplitude, and a central difference method to compute joint velocities and accelerations from joint positions to reduce any distortion in the data. Specifically, we selected a 4-th order butterworth filter with cutoff frequency of 10Hz. The identification of the latent variable model is performed using only torque and joints position measurements. From these, the joint velocities, accelerations and latent states, which are needed to evaluate the dynamic model, are estimated through the particle filter.

For implementation and identification details of the Stribeck, Lugre, and GMS models, we follow established best practices as described in \cite{stribeck}, \cite{Lugre}, and \cite{GMS}, respectively. In the case of the data-driven models, a comprehensive hyperparameter search was conducted to optimize performance. The final NN, RNN and LVM models were trained using the Adam optimizer, with a learning rate of 0.001. The batch sizes were set to 64 for the NN and 128 for the RNN. The NN architecture consists of two hidden layers, each with 32 nodes. The RNN architecture comprises three layers with 32 nodes and a hidden state size of 32. Both models apply the ReLU activation function. 

The LVM architecture comprises a neural network with a single hidden layer of 32 nodes for the latent dynamics function and a neural network with two hidden layers of 32 nodes for the friction function, both  with the Mish activation function applied. The optimal dimension of the latent state $z$, was determined to be 2. During training, 200 particles were used for the SMC methods.

\subsection{Dynamic simulation}
Since the robot manufacturer does not disclose the inertial and dynamic parameters, the validation of the proposed method was carried out by assessing the open-loop prediction performance. These open-loop simulations were performed by applying the motor torques measured for the validation reference trajectory and comparing the predicted joint positions and velocities with the corresponding measured values. Given that we do not have access to the internal controller responsible for tracking the reference signals, we chose to not include control dynamics in the robotic model and to directly apply the measured motor torques. 

\begin{figure}[h]
    \centering
    \vspace{-10pt}
    \includegraphics[width=\columnwidth]{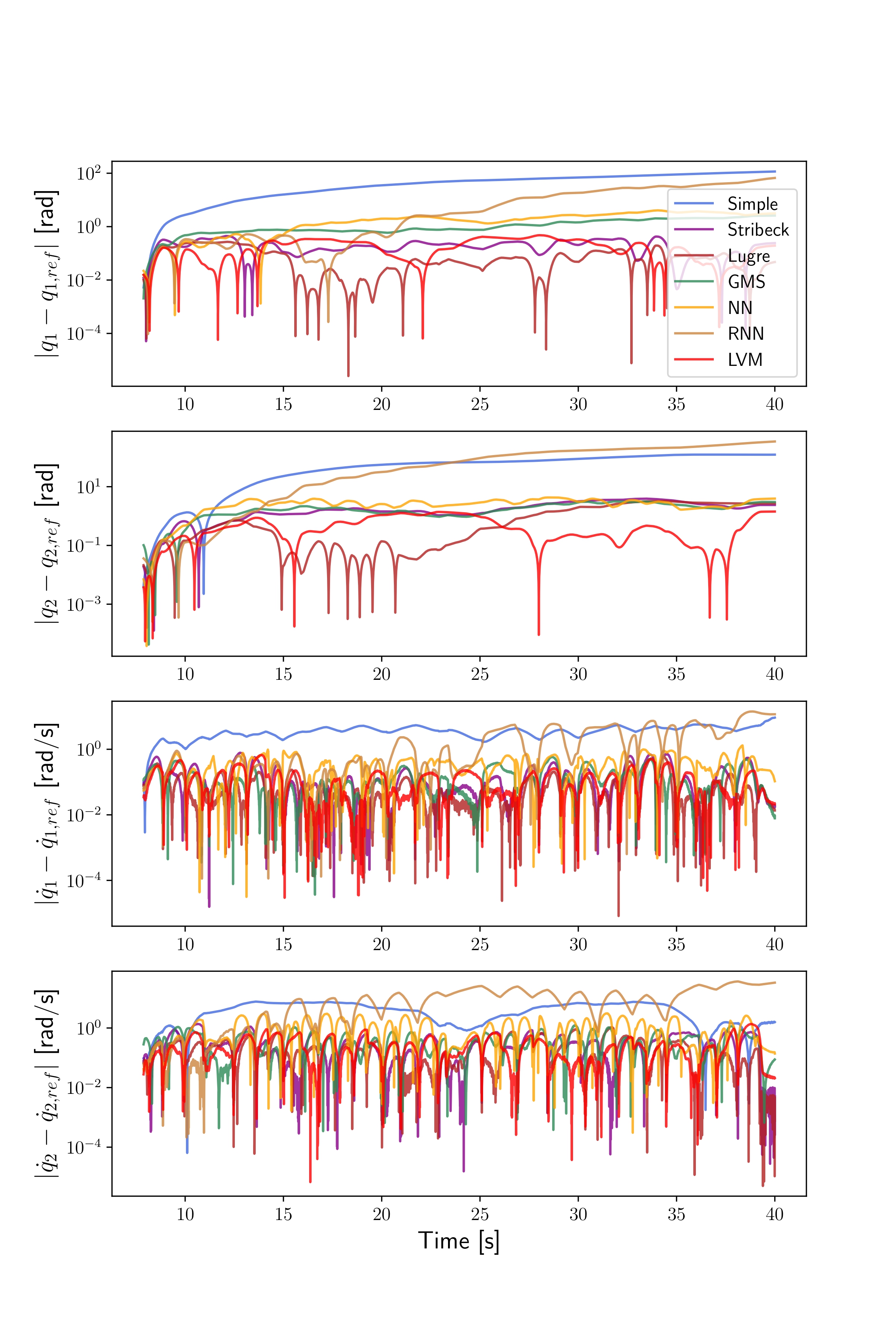}
    \vspace{-40pt}
    \caption{Absolute errors of the open-loop estimations of the different models with respect to the reference signal.}
    \label{fig:openloop}
\end{figure}

The open-loop prediction results for the complete test trajectory, as well as the first 10 seconds, are quantitatively summarized in Table \ref{table:openloop} using the Mean Squared Error (MSE) and Mean Absolute Error (MAE) metrics. Fig. \ref{fig:openloop} depicts the absolute error of the predicted open-loop signals of the different models w.r.t. the reference validation trajectory. Note to logarithmic scale on the vertical axes. The physics-based benchmark models, i.e.  the Lugre and GMS model, deliver robust open-loop predictions over the complete trajectory, but fail to accurately capture all the subtilities in the dynamics, for example around direction reversals. The data-driven benchmark models, i.e. the fully connected NN and RNN, are more accurate at the beginning of the test trajectory. However, as the trajectory progresses, the accumulated error increases to the extent that the inputs to these models encounter values outside the state-space covered in the training data, leading to unstable outputs and exploding values in the predictions. For the simple model and the RNN, the accumulated errors became so significant that their MAE and MSE values were omitted from the analysis of long prediction horizons. The proposed data-driven latent variable model outperforms all benchmark models in both accuracy and robustness. The dynamics identified by the latent state variables helps improving the stability and accuracy of the predictions.

\begin{table}[t]
\centering
    \caption{Open-loop prediction of the different models. Quantative results in terms of MAE en MSE.}
    \begin{tabular}{c c c c c}
        \toprule
        \textbf{Method} & \multicolumn{2}{c}{\textbf{$10s$ interval}} & \multicolumn{2}{c}{\textbf{Complete trajectory}}  \\
            \cmidrule(lr){2-3} \cmidrule(lr){4-5} 
             & \textbf{MSE}  & \textbf{MAE} & \textbf{MSE}  & \textbf{MAE}\\
         \midrule
         Simple & $1.97$ & $0.96$ & $-$ & $-$ \\
         \midrule
         Stribeck & $0.14$ & $0.25$ & $1.15$ & $0.61$ \\
         \midrule
         Lugre & $0.027$ & $0.12$ & $0.66$ & $0.42$ \\
         \midrule
         GMS & $0.15$ & $0.28$ & $1.43$ & $0.84$ \\
         \midrule
         Fully connected NN & $0.21$ & $0.26$ & $2.29$ & $1.07$ \\
         \midrule
         RNN & $0.039$ & $0.13$ & $-$ & $-$ \\
         \midrule
         LVM (ours) & $0.024$ & $0.11$ & $0.11$ & $0.22$ \\
         \bottomrule
    \end{tabular}
    \label{table:openloop}
\end{table}

\subsection{Identified friction characteristic}
\begin{figure}[h]
    \centering
    \begin{subfigure}{0.49\columnwidth}
        \centering
        \includegraphics[width=\textwidth]{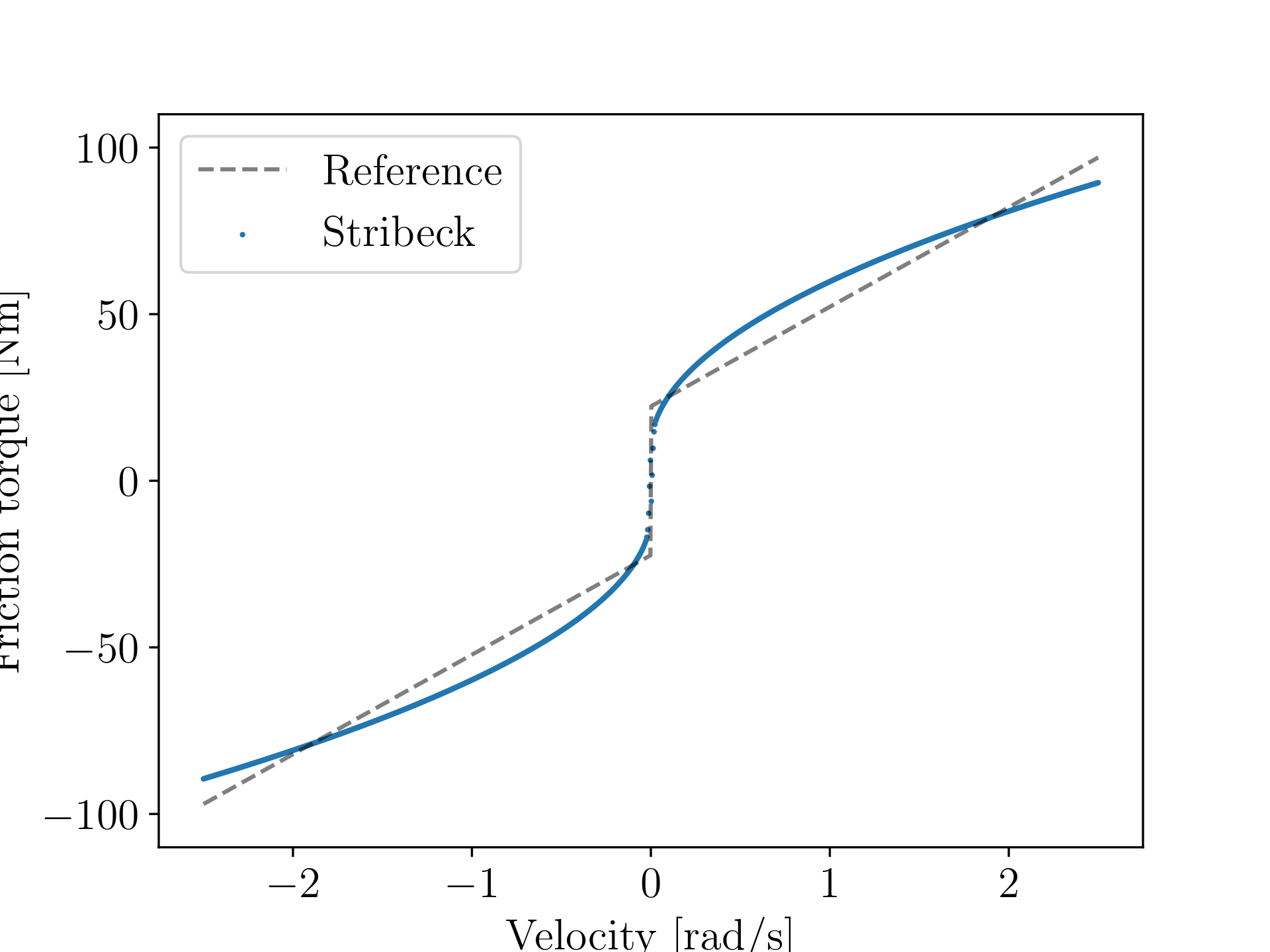}
        \caption{Stribeck}
    \end{subfigure}
    \begin{subfigure}{0.49\columnwidth}
        \centering
        \includegraphics[width=\textwidth]{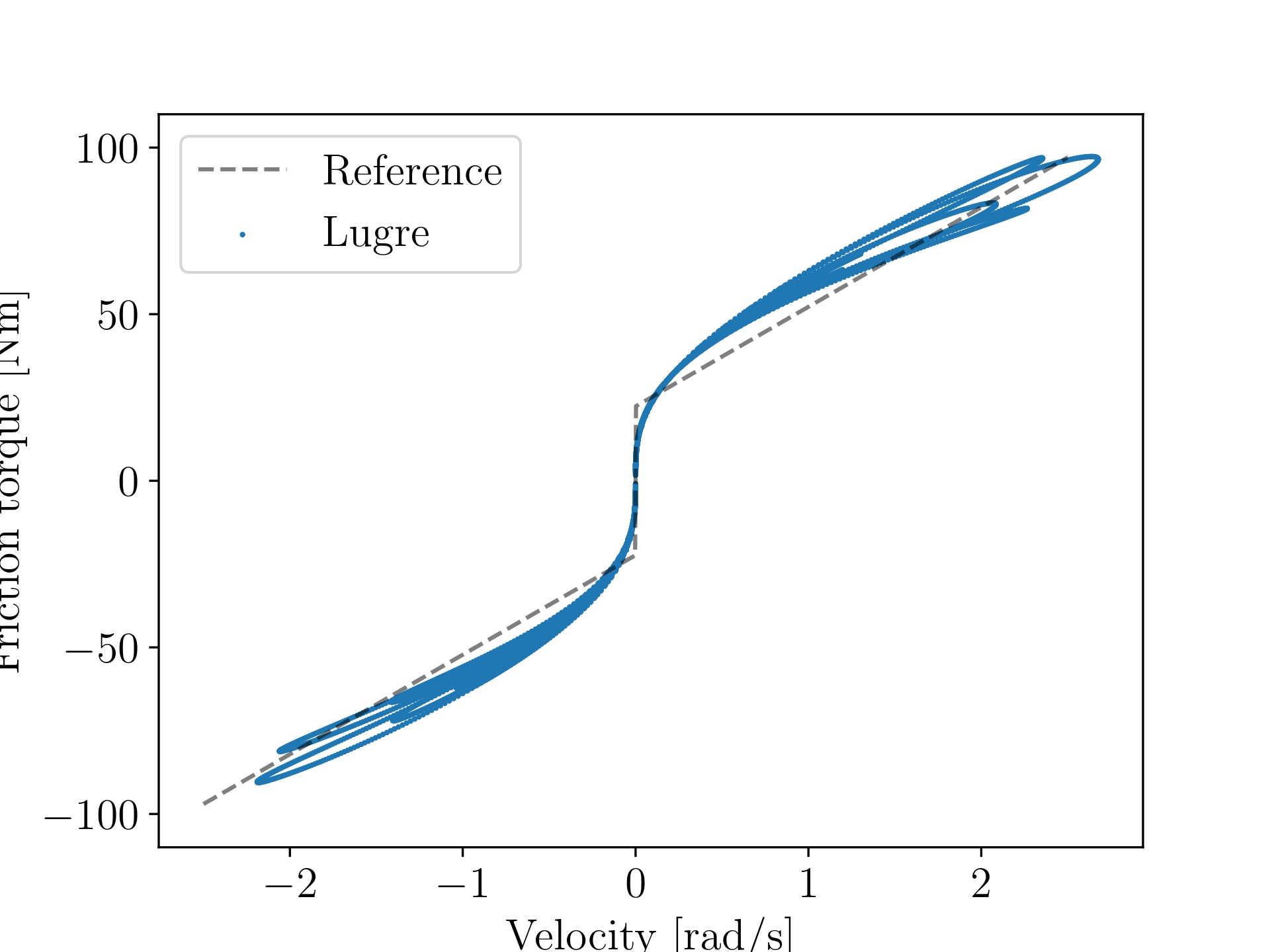}
        \caption{Lugre}
    \end{subfigure}
    \begin{subfigure}{0.49\columnwidth}
        \centering
        \includegraphics[width=\textwidth]{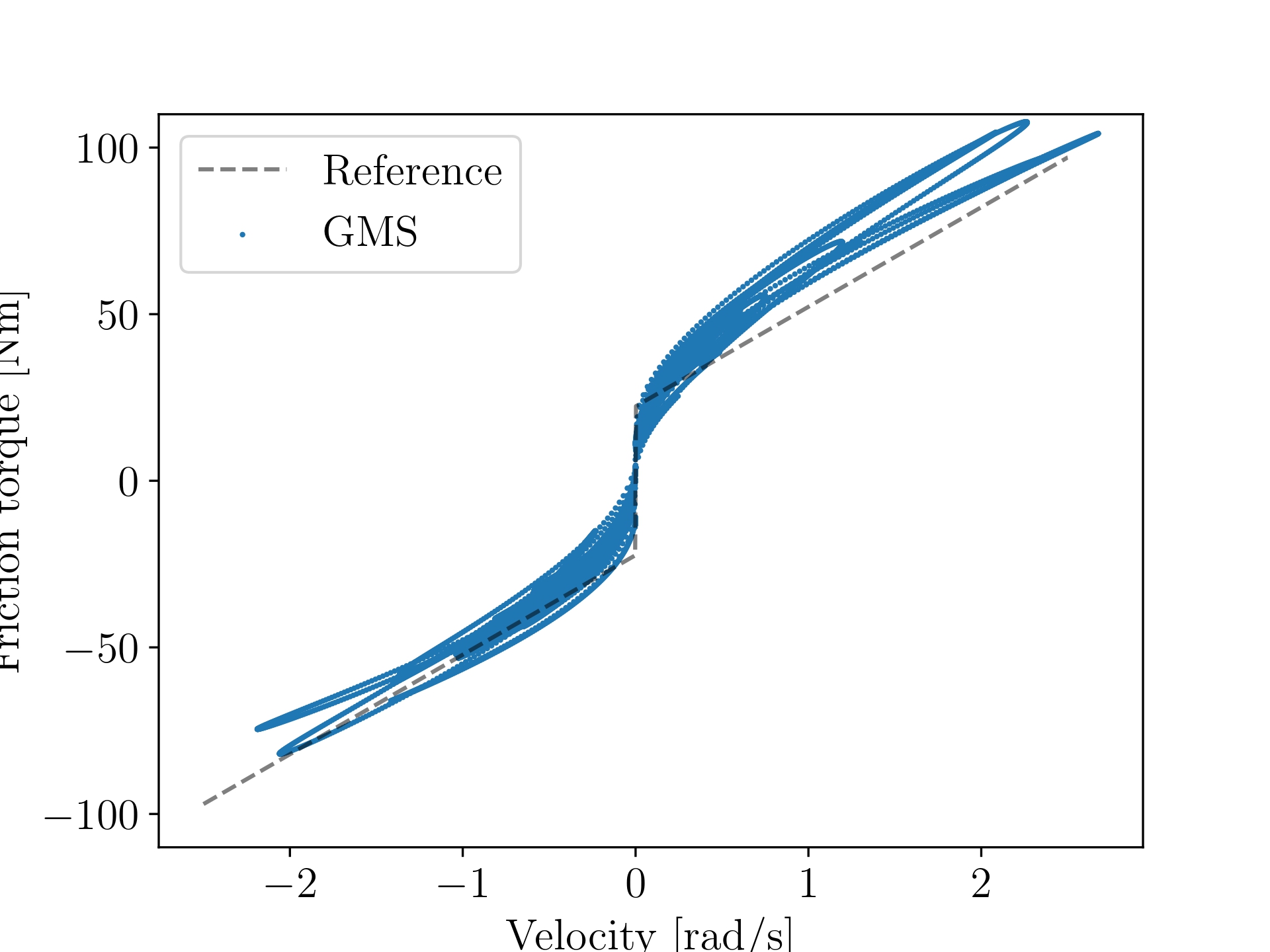}
        \caption{GMS}
    \end{subfigure}
    \begin{subfigure}{0.49\columnwidth}
        \centering
        \includegraphics[width=\textwidth]{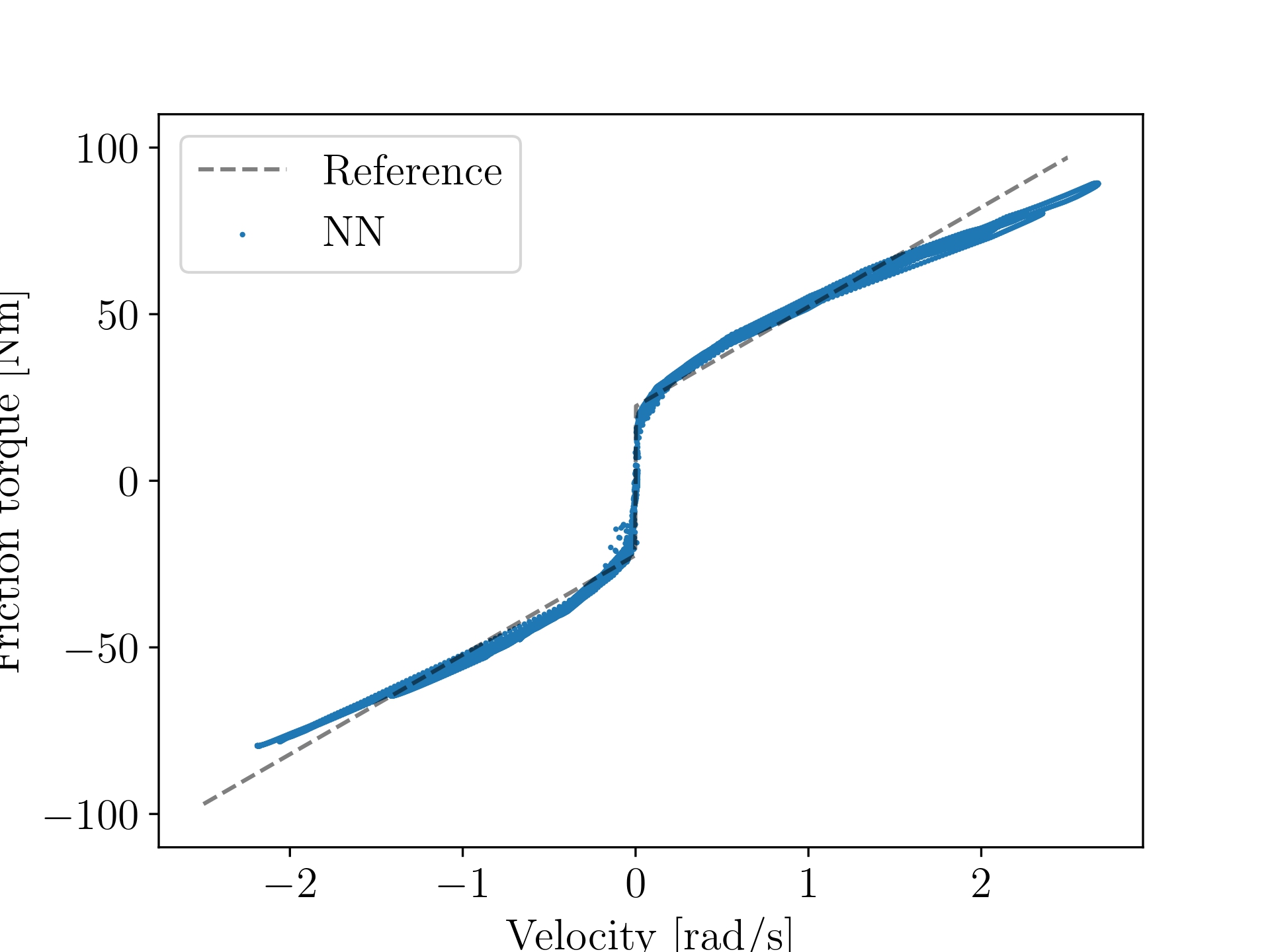}
        \caption{Fully connected NN}
    \end{subfigure}
    \begin{subfigure}{0.49\columnwidth}
        \centering
        \includegraphics[width=\textwidth]{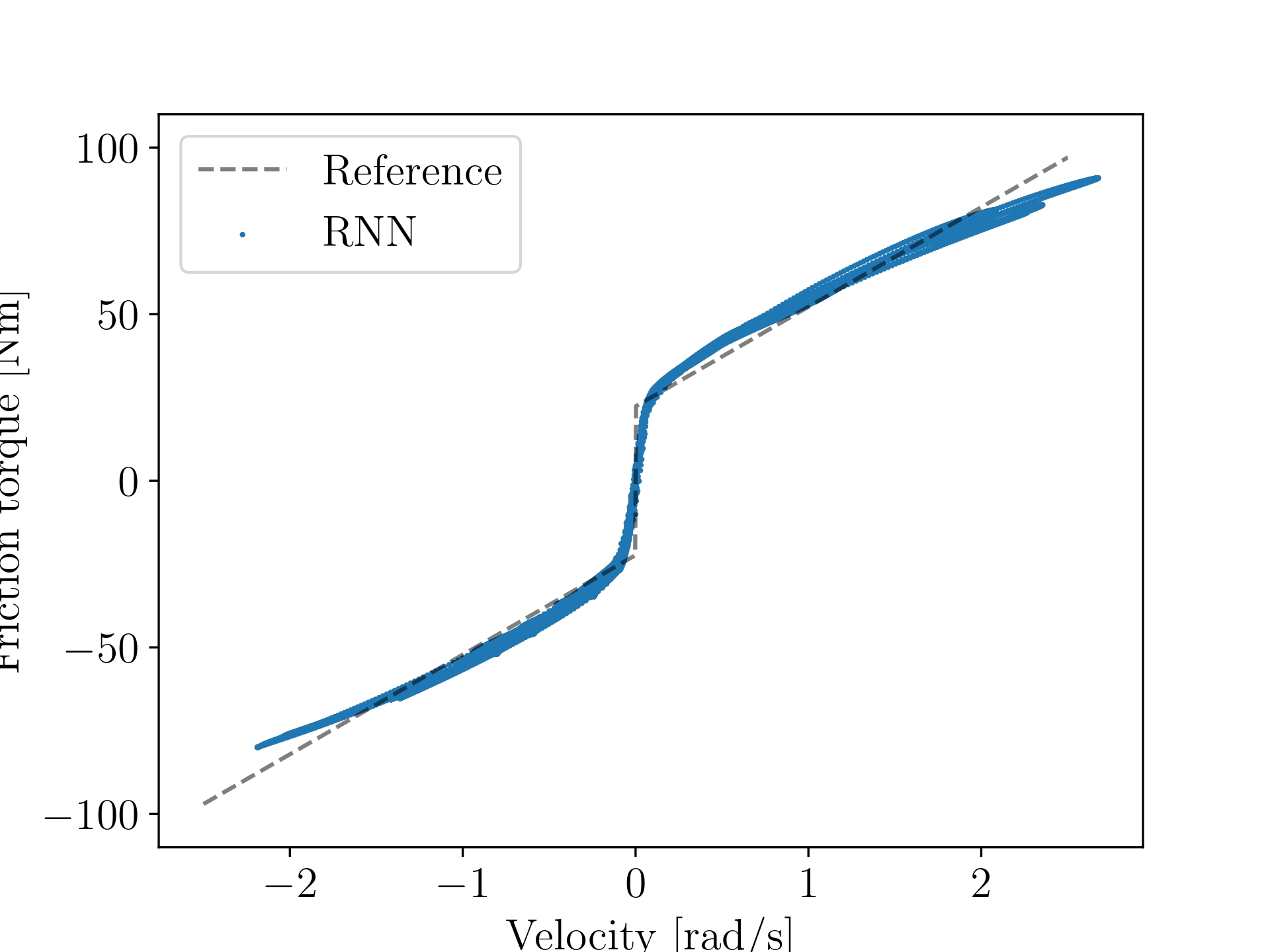}
        \caption{RNN}
    \end{subfigure}
    \begin{subfigure}{0.49\columnwidth}
        \centering
        \includegraphics[width=\textwidth]{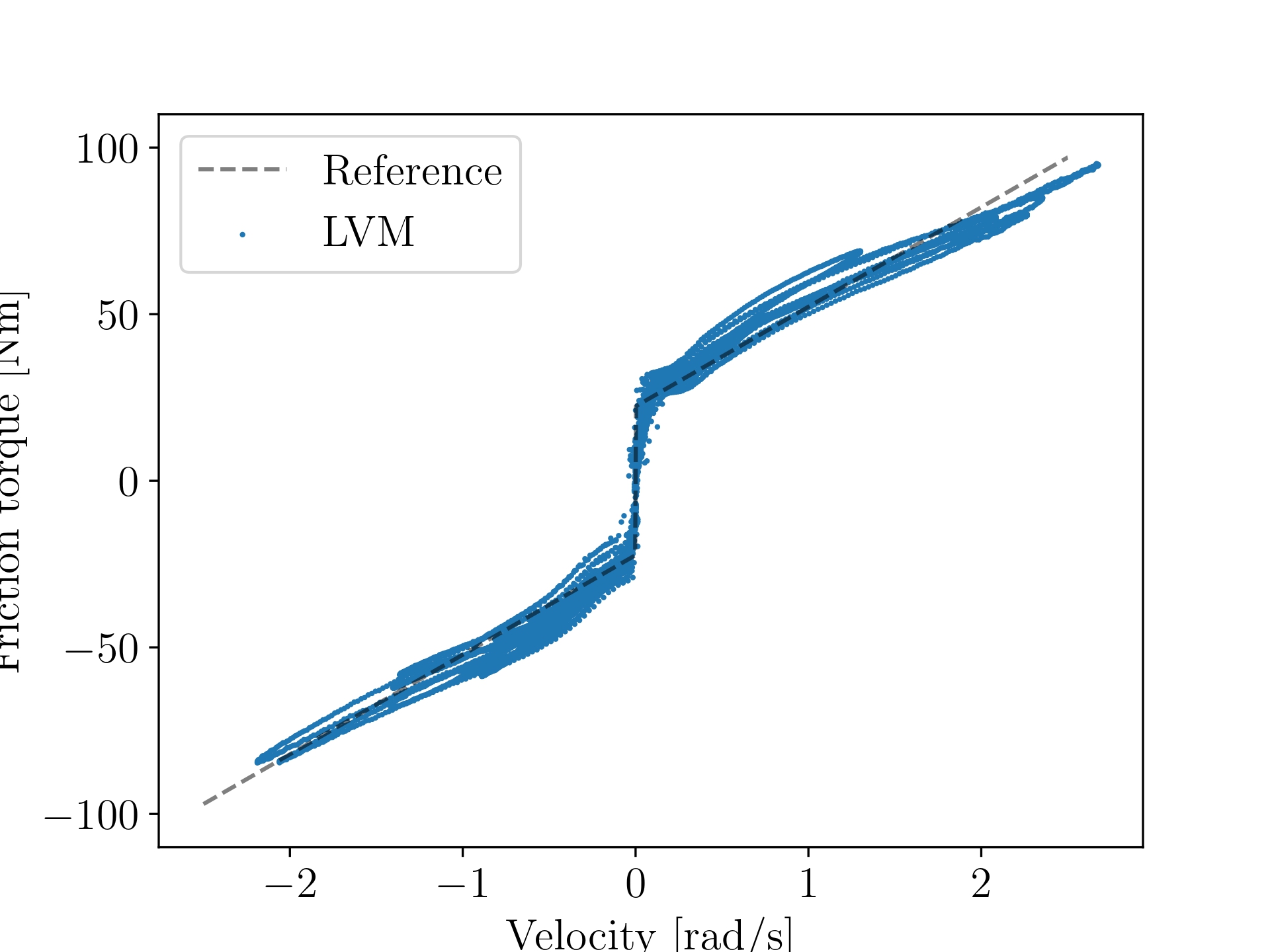}
        \caption{LVM (ours)}
    \end{subfigure}
    \caption{The estimated friction characteristics for the different models of joint 1.}
    \label{fig:fricchar}
\end{figure}

For completeness, the friction characteristics identified by the different models for Joint 1 are shown in Fig. \ref{fig:fricchar}. These friction characteristics were derived by evaluating the models on the measured joint positions, velocities and motor torques. Due to the lack of joint torque sensors in the KUKA KR6 R700, no direct ground truth for the friction characteristics is available, precluding a quantitative comparison. Nonetheless, these results are provided to give a qualitative perspective on the friction behavior and offer insight into the order of magnitude of the estimated friction effects. For reference, the friction characteristic identified by the simple model is included, enabling an indirect comparison between the models.

The friction characteristic of the proposed LVM displays more nuanced dynamics in the low velocity range, compared to other models. Static friction, or stiction, is well captured near zero velocity, as indicated by the sharp transition around 0 rad/s, where friction changes from negative to positive values. In the low-velocity range, the model captures the Stribeck effect, characterized by a reduction in friction with increasing velocity before the transition to a velocity-strengthening regime. As velocity increases further, the model identifies the dominant viscous friction, where friction force increases more linearly with velocity. Additionally, the identified model exhibits clear hysteresis loops, indicating the presence of memory effects in the friction dynamics, reflecting the path-dependent nature of friction. These features highlight the model's capability to capture the complex frictional behavior across different velocity regimes.

\section{Conclusion}\label{sec:conclusion}
This paper proposes probabilistic LVMs for friction modelling in robot joints. Data-driven modelling techniques, here neural networks, are inserted in the dynamic model and serve as a highly flexible parametrization to identify the nonlinear friction behaviour in robotic joints. The system state is augmented by latent variables to account for unmodeled and unknown underlying phenomena influencing the robot dynamics. The friction characteristic and latent dynamics are learned, simultaneously with the other base inertial parameters describing the lumped parameter model of the robot dynamics, directly from noisy sensor data. 
The inherently stochastic and unsupervised nature of the identification problem is addressed by framing it as a probabilistic learning problem. A Maximum Likelihood Estimate of the model parameters is obtained using the Expectation-Maximization algorithm in conjuction Sequential Monte Carlo techniques. This approach also relaxes the demands on the Design of Experiments and eliminates the need for pre-processing of the training data. Experimental validation on the KUKA KR6 R700 shows that the proposed methodology can accurately identify the dynamic model. This, however, comes at a cost of increased computational complexity compared to conventional modelling methods. Further research should be conducted to validate the proposed methodoly for full 6 or 7 degree-of-freedom robotic systems. Additionally, future work should explore and identify structures within the latent dynamics that could enhance the proposed parameterization, thereby reducing both identification and computational complexity.

\section*{Acknowledgements}
This work was supported by the Flanders Make project QUASIMO and the Flanders AI Research Programme.

\end{document}